\documentclass[letterpaper, 10 pt, conference]{ieeeconf}  

\IEEEoverridecommandlockouts                              

\DeclareRobustCommand{\rchi}{{\mathpalette\irchi\relax}}
\newcommand{\irchi}[2]{\raisebox{\depth}{$#1\chi$}} 
\usepackage{amsmath}
\usepackage{amssymb}
\usepackage{amsmath}
\usepackage{caption}
\usepackage{subcaption}
\DeclareMathOperator*{\argmax}{arg\,max}

\usepackage{subfiles} 
\usepackage{graphicx}
\graphicspath{ {./Figures/} }
\usepackage{algorithm} 
\usepackage{algpseudocode} 
\usepackage{afterpage,lipsum}
\usepackage{lipsum}
\usepackage{graphicx,dblfloatfix}

\usepackage{hyperref}
\usepackage{cite}

\usepackage{svg}
\algnewcommand{\Input}{\textbf{Input}}
\algnewcommand{\Initialize}{\textbf{Initialize}}
\title{\LARGE \bf
Active Tapping via Gaussian Process for Efficient Unknown Object Surface Reconstruction
}

\author{Su Sun and Byung-Cheol Min
\thanks{The authors are with SMART Lab, Department of Computer and Information Technology, Purdue University, West Lafayette, IN 47907, USA {\tt\small sun931@purdue.edu | minb@purdue.edu}}%
}

\begin{document}

\maketitle
\thispagestyle{empty}
\pagestyle{empty}

\begin{abstract}
Object surface reconstruction brings essential benefits to robot grasping, object recognition, and object manipulation. When measuring the surface distribution of an unknown object by tapping, the greatest challenge is to select tapping positions efficiently and accurately without prior knowledge of object region. Given a searching range, we propose an active exploration method, to efficiently and intelligently guide the tapping to learn the object surface without exhaustive and unnecessary off-surface tapping. We analyze the performance of our approach in modeling object surfaces within an exploration range larger than the object using a robot arm equipped with an end-of-arm tapping tool to execute tapping motions. Experimental results show that the approach successfully models the surface of unknown objects with a relative 59\% improvement in the proportion of necessary taps among all taps compared with state-of-art performance.

\end{abstract}
\section{Introduction}
Surface reconstruction of unknown objects allows various applications such as robot grasping\cite{5354345}, object recognition, and object manipulation\cite{8594009}. Tactile sensors, as a common tool for shape recognition, provide intuitive information of objects by direct interaction on the object's surface\cite{8468088}. Although tactile sensors reduce the occlusion problem existing in vision sensors, they suffer from sensory noise and require lengthy contact duration with the surface \cite{9341346}. Within touch methods presented in the literature, random search and grid search methods are most common in unknown object shape recognition without prior information gained from the object. Intuitively, the main drawbacks of such methods are that they are inefficient due to a large amount of contact, less robustness for a mandatory resolution, and also require pre-defined searching regions.

In this paper, we propose an active tapping method for unknown object surface reconstruction. The goal of this study is to learn the object surface shape using tapping motions without prior knowledge, and maximize valid on-surface taps to increase accuracy and efficiency of surface reconstruction. In order to acquire assessment of object surfaces, the tapping motion is conducted by Baxter robot arm, during which the force on an end-of-arm tool is detected by the force sensor to recognize touches on objects. More specifically, three dimensional positions of tapping on object's surface are determined by forward kinematics of the robot arm in world coordinates, where the height values are recorded when contacts with object surface are recognized. A probabilistic representation of the object surface is generated by Gaussian Process Regression (GPR) in an explicit way. An active searching approach is proposed to efficiently and intelligently guide the tapping positions to reduce the uncertainty in the regression function. The main contributions of our work are as follows: 
\begin{itemize}

\item We use tapping motion to gain object assessment by force data on the robot arm's end effector.
\item We use Gaussian Processes as surface potential, with an active searching method to suggest a next point to tap and avoid exhaustive exploration.
\item We do not require prior-knowledge of object regions. Accuracy and efficiency of surface reconstruction is prompted by maximizing on-surface tapping and minimizing unnecessary off-surface tapping.

\end{itemize}

\begin{figure}[t]
  \centering\includegraphics[width=.85\linewidth]{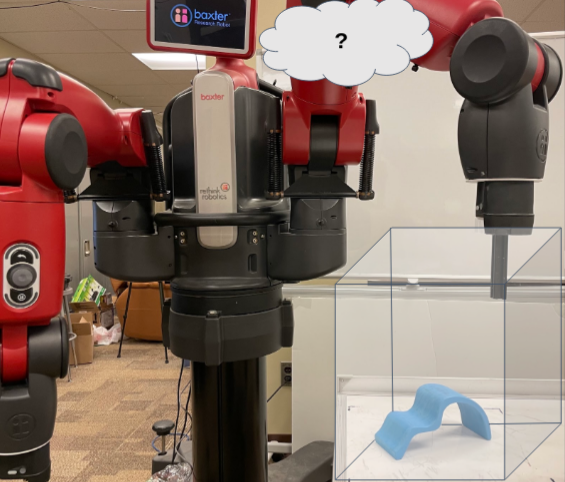}
  \caption{Conceptual illustration of the robot estimating object surface by an active tapping via Gaussian process in an unknown searching space.}
    \label{fig:fig1}
    \vspace{-3mm}
\end{figure}

\section{Active Tapping with Gaussian Process for Object Surface Reconstruction}

\subsection{Problem Statement}
The main problem of this work is to efficiently search for tapping positions without knowing object edge and region. We apply GPR to represent the uncertainty of surface estimation and an acquisition function to guide the tapping points to minimize the uncertainty of target function. The surface representation model based on GPR is detailed in Section \ref{GP}. In consideration of lacking the prior information of object region, there exists a great uncertainty in off-surface space in the searching area, where less tapping should be executed. In order to guide the tapping to focus on object surfaces to enhance accuracy, we use a weight function to cumulatively place weights on each tapping point. To be more specific, unprofitable tapping happening within off-surface space, which is called unprofitable areas, will result in continuously reduced weights in adjacent areas in each iteration. Intuitively, after the small number of unprofitable tapping is conducted, most of next tapping positions are concentrated within on-surface space. The active searching function is elaborated in Section \ref{as}.


 \subsection{Gaussian Processes for Surface Estimation }\label{GP} 
 We define the problem as a general regression problem. Given an exploration domain $\rchi\rightarrow \mathbb{R}^d$, an unknown surface function $f: \rchi$, and a set of observations of the function $D ={(\mathbf{x}_i,f(\mathbf{x}_i))}$ where $\mathbf{x}_i$ is the tapping position in exploration space, and $f(\mathbf{x}_i)$ is the tapping result including surface height estimation, our goal is to predict the value of the function $\mathbf{f_*} = f(\mathbf{X}_*)$ at positions $\mathbf{X}_*$, thus building the surface model.
 
The GP distribution is defined as $p(f) = \mathcal{GP}(f;\mu,K)$ where $\mu$ is the mean function, and $K$ is associated kernel function. Given a finite set of points $\mathbf{X} \subseteq \rchi$, the GP prior on $f$ is given by a joint distribution on $\mathbf{f} = f(\mathbf{X})$:
\begin{equation}
 p(\mathbf{f}|\mathbf{X}) = \mathcal{N}(\mathbf{f}; \mu(\mathbf{X}),\mathbf{K}),
\end{equation} where $\mathbf{K} = k(\mathbf{X}, \mathbf{X})$ is the covariance matrix for all point pairs in $\mathbf{X}$. The GP prior mean $\mu(\mathbf{X})$ is set to zero.

To predict the values of $f$ at $\mathbf{X}_*$, a joint distribution of $\mathbf{X}$ and $\mathbf{f_*}$ is written as:
\begin{equation}
p(\mathbf{f},\mathbf{f}_*) = \mathcal{N}(\begin{bmatrix}
\mathbf{f}\\\mathbf{f}_* \end{bmatrix}; \begin{bmatrix}
\mu(\mathbf{X})\\ \mu(\mathbf{X}_*) \end{bmatrix}, \begin{bmatrix}
\mathbf{K}, \mathbf{K}_*\\ \mathbf{K}_*, \mathbf{K}_*{}_*\end{bmatrix}),
\end{equation} where $\mathbf{K}_* = k(\mathbf{X}, \mathbf{X}_*)$, and $\mathbf{K}_*{}_* = k(\mathbf{X}_*, \mathbf{X}_*)$.

Using the Gaussian conditioning, the posterior distribution can be computed as:

\begin{equation}
\!
\begin{aligned}
&p(\mathbf{f}_*|\mathbf{X}_*, \mathbf{X}, \mathbf{f}) = \mathcal{N}(\mathbf{f}_*|\mu(\mathbf{X}_*), \Sigma(\mathbf{X}_*))\\
&\mu(\mathbf{X}_*) = \mu(\mathbf{X}_*) + \mathbf{K}_*^T\mathbf{K}^{-1}(\mathbf{f}-\mu(\mathbf{X}))\\
&\Sigma(\mathbf{X}_*) = \mathbf{K}_*{}_* - \mathbf{K}_*^T\mathbf{K}^{-1}\mathbf{K}_*.
\end{aligned}\label{eq:label3}
\end{equation}

We utilize the radial-basis function (RBF) kernel based on our assumption that nearby points share close function values in surface distribution in this study. The RBF kernel is given as:
\begin{equation}
k(\mathbf{x}_i,\mathbf{x}_j) = \exp(-\frac{1}{2}(\frac{d(\mathbf{x}_i, \mathbf{x}_j)}{\sigma})^2)
\end{equation} where $d$ is the Euclidean distance between two points, and $\sigma$ is the hyperparameter. 

For each point within the posterior distribution, the mean which encodes the predictions at corresponding position and the variance which represents the confidence level of current prediction are given by GPR. Based on the definition of RBF kernel, the variance at predicted points decreases as the distance to the observed points decreases. At each iteration $i$ when tapping is done, a new point $(\mathbf{x}_i,f(\mathbf{x}_i))$ is fit into the GP, and posteriors are updated to better approximate the unknown surface function.
\begin{figure*}
    \centering
    \begin{minipage}{.2\linewidth}
            \begin{subfigure}[t]{.9\linewidth}
                \includegraphics[width=\textwidth]{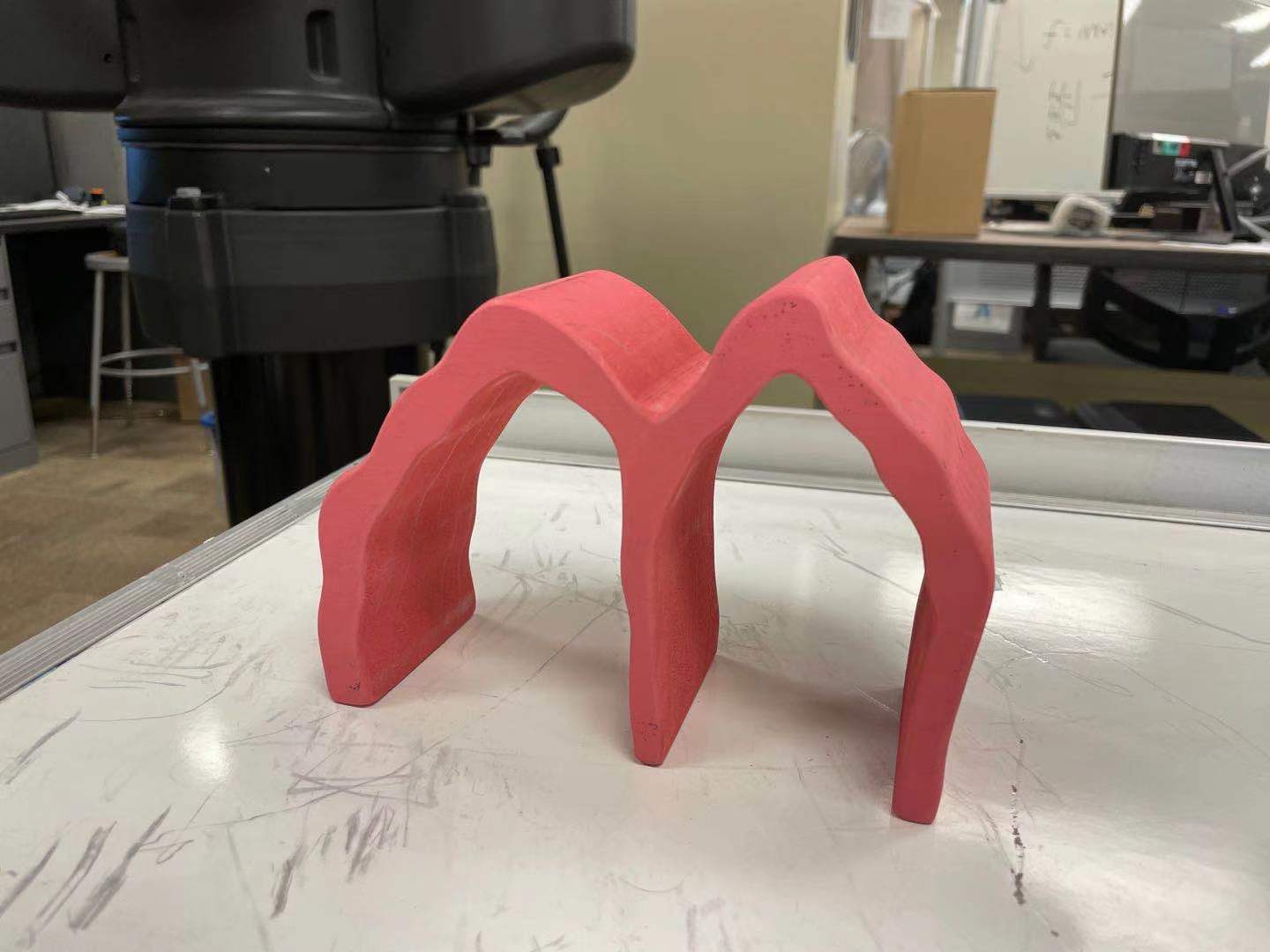}
                \caption{Object I}
                \label{fig:weather_activity}
            \end{subfigure}
            \begin{subfigure}[t]{.9\linewidth}
                \includegraphics[width=\textwidth]{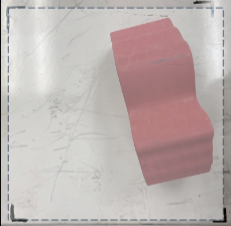}
                \caption{searching space}
                \label{fig:weather_activity}
            \end{subfigure}
        \end{minipage}
    \begin{minipage}{.79\linewidth}
        \begin{subfigure}[t]{.24\linewidth}
            \includegraphics[width=\textwidth,height=0.11\textheight]{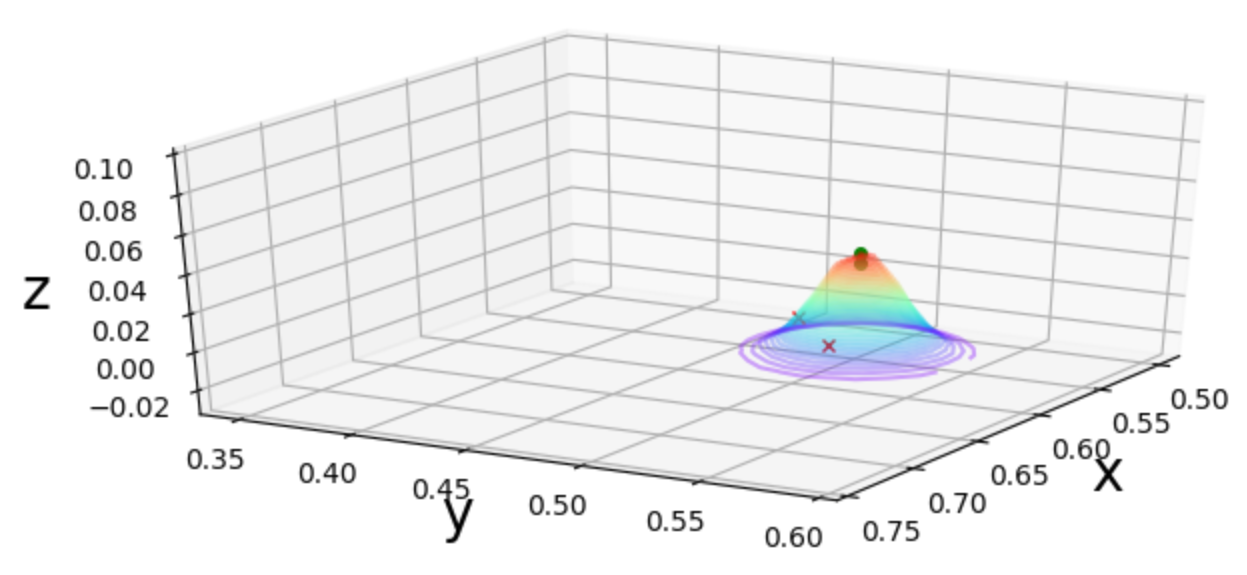}
            \label{fig:weather_filter1}
        \end{subfigure} 
        \begin{subfigure}[t]{.24\linewidth}
            \includegraphics[width=\textwidth,height=0.11\textheight]{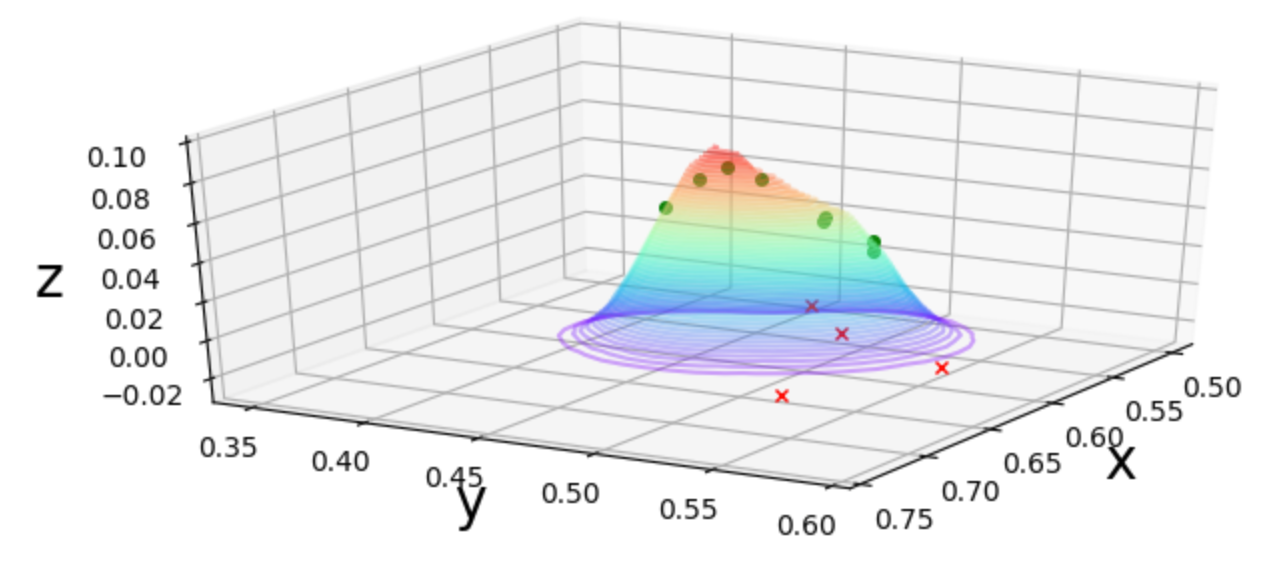}
            \label{fig:weather_filter1}
        \end{subfigure} 
         \begin{subfigure}[t]{.24\linewidth}
            \includegraphics[width=\textwidth,height=0.11\textheight]{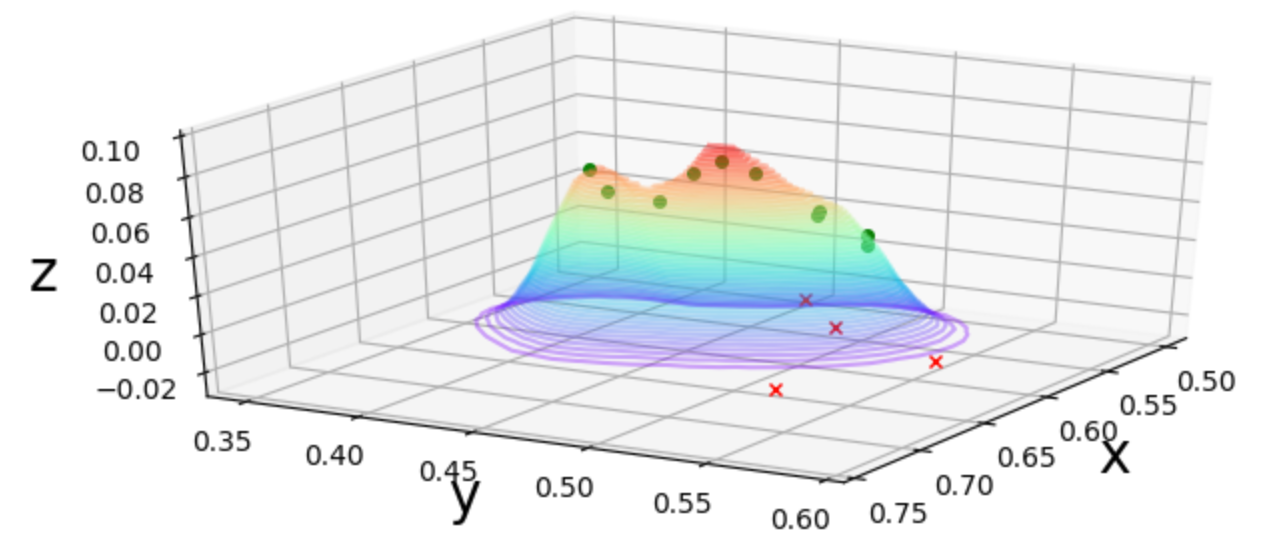}
            \label{fig:weather_filter1}
        \end{subfigure} 
         \begin{subfigure}[t]{.24\linewidth}
            \includegraphics[width=\textwidth,height=0.11\textheight]{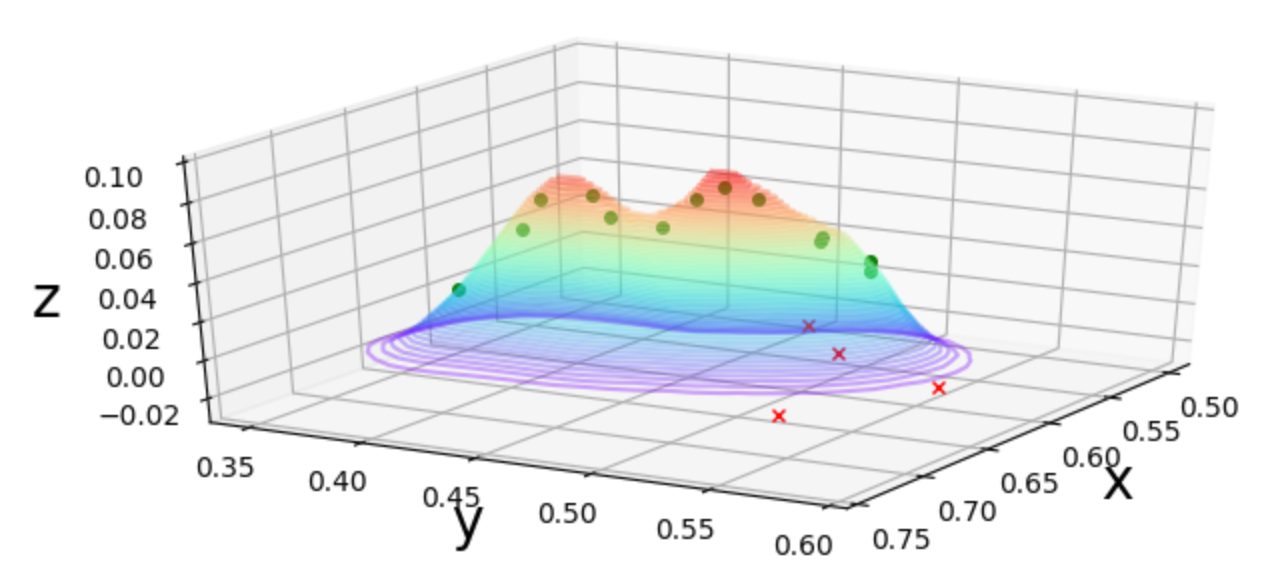}
            \label{fig:weather_filter1}
        \end{subfigure} \\
        \medskip
         \begin{subfigure}[b]{.24\linewidth}
            \includegraphics[width=\textwidth,height=0.11\textheight]{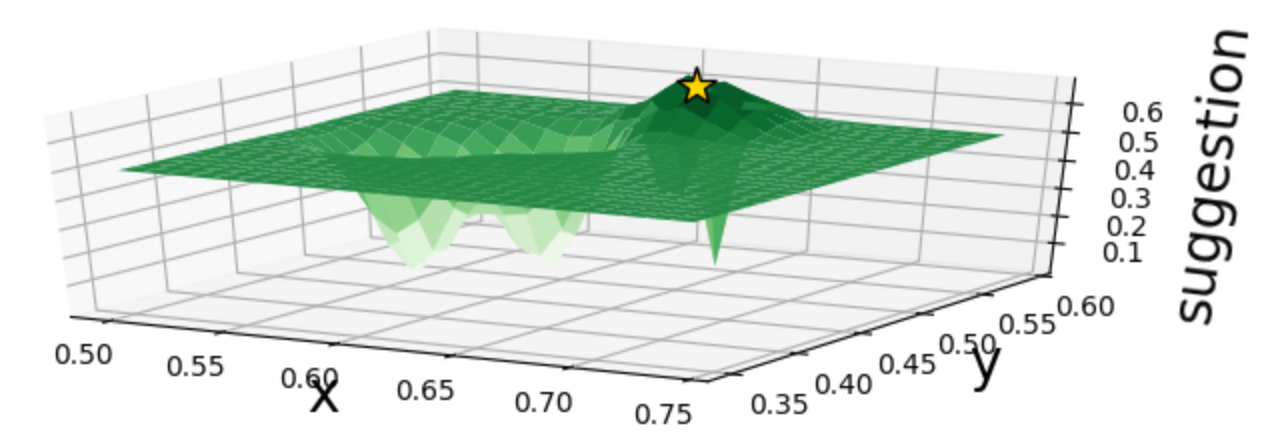}
            \caption{Iteration 1}
            \label{fig:weather_filter1}
        \end{subfigure} 
        \medskip
        \begin{subfigure}[b]{.24\linewidth}
            \includegraphics[width=\textwidth,height=0.11\textheight]{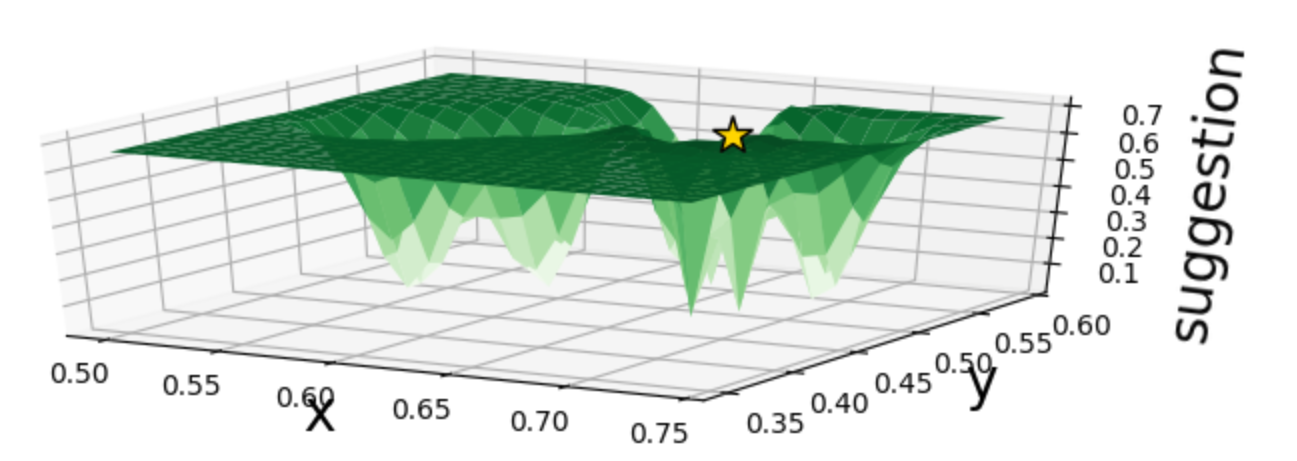}
            \caption{Iteration 8}
            \label{fig:weather_filter1}
        \end{subfigure} 
        \medskip
         \begin{subfigure}[b]{.24\linewidth}
            \includegraphics[width=\textwidth,height=0.11\textheight]{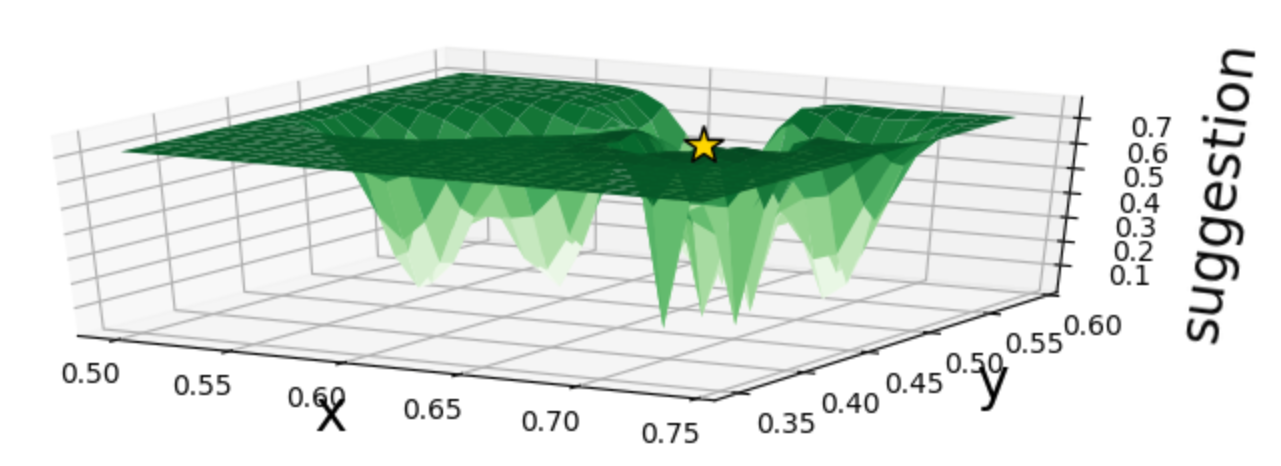}
            \caption{Iteration 10}
            \label{fig:weather_filter1}
        \end{subfigure} 
        \medskip
         \begin{subfigure}[b]{.24\linewidth}
            \includegraphics[width=\textwidth,height=0.11\textheight]{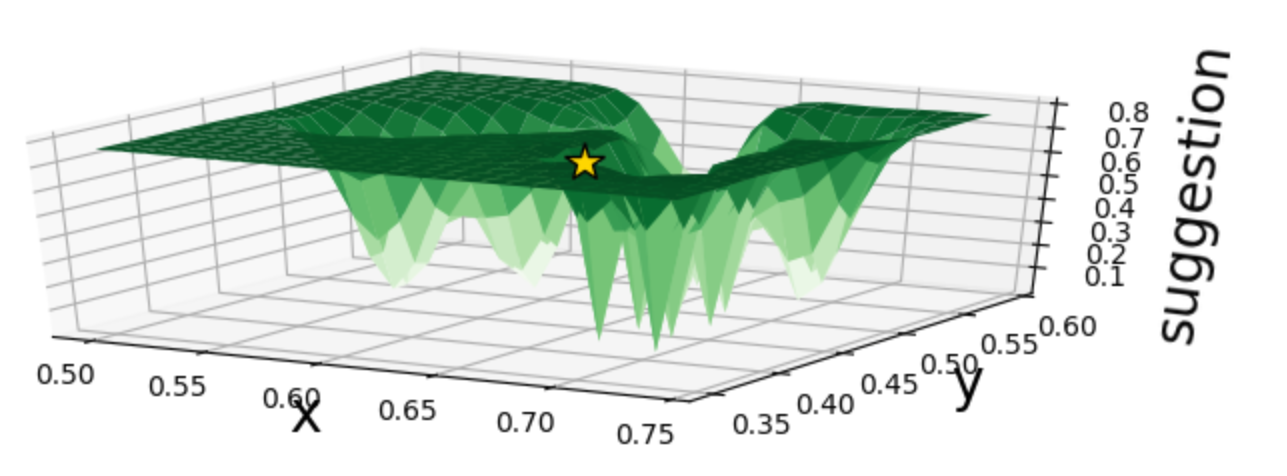}
            \caption{Iteration 14}
            \label{fig:weather_filter1}
        \end{subfigure} 
    \end{minipage}
    \vspace{-2mm}
    \caption{(a) a toy block with wave shape as object I; (b) given searching space; and (c-f) GP model (the first row) and corresponding suggestion function (the second row) at four non-consecutive taps. The star marker on suggestion function represents the next suggested point. At the first iterations, three tapping points generated randomly are executed as input of GP model. The hyperparameter used is RBF kernel with $\sigma^2 = 0.017$. \textit{Please zoom-in on (c-f) for details.}}
\label{fig: result}
\end{figure*}
\subsection{Active Exploration Function }\label{as}
 As every tapping trail is costly, we intent to avoid a brute-force tapping in search space and concentrate tapping on object surface. In order to build surface model accurately and efficiently, we exploit an active exploration function which suggests the tapping positions at each iteration. The exploration function is interpreted as a joint outcome of an uncertainty function and a weight function, to respectively minimize the uncertainty of surface model estimation and place weights on possibly suggested points. The uncertainty function and the weight function are accordingly elaborated as follows.
 
 \noindent\textbf{Uncertainty Function.~}
 Based on attributes of the GP posterior distribution, with $n$ posteriors, the uncertainty function is given as:
 \begin{equation}
 u(\mathbf{X}; D_n) : \mathbf{X}\rightarrow \rchi = \Sigma(\mathbf{X}).
\end{equation}
At $t_{\text{th}}$ iteration, the uncertainty function suggests tapping point $\mathbf{x}_t$ by:
 \begin{equation}
 \mathbf{x}_t = \argmax_\mathbf{x} u.
\end{equation}

Given that the variance at observed points is zero, uncertainty of each prediction decreases as it gets closer to observations. The advantage of using the uncertainty function is that the predicted point with the highest uncertainty will be sampled to reduce the uncertainties of the whole surrogate function, thus approaching the true function by a small number of essential samples. 
 
 \noindent\textbf{Weight Function.~}
As the uncertainty function guides tapping to positions with the highest uncertainty, there still exit a drawback in using the uncertainty function alone. When the object edge is unknown, uncertainties will not only exist in on-surface space, but also off-surface space. As redundant off-surface samples will not enhance the speed and accuracy of surface model, exhaustive searches in those areas need to be minimized. In order to avoid redundant sampling, we employ an additional GPR to model the possibility of unprofitable tapping at each point. We model the possibility at each point as:
 \begin{equation}
f_w(\mathbf{x}): \mathbf{x}\rightarrow\mathbb{R}^d  = 
     \begin{cases}
       1, &\text{if } \mathbf{x} \text{ on surface} \\
       0, &\text{if } \mathbf{x} \text{ off surface}. \\
     \end{cases}
\end{equation}

$H(\mathbf{x}_i,f_w(\mathbf{x}_i))$, as weights of explored points, is fit into a GP model given as:
\begin{equation}
p(\mathbf{f_w}_*|\mathbf{X}_*, \mathbf{X}, \mathbf{f_w}) = \mathcal{N}(\mathbf{f_w}_*|\mu_w(\mathbf{X}_*), \Sigma_w(\mathbf{X}_*))
\end{equation} with the RBF kernel.
The GP model, therefore, predicts the probability of each point within the domain being on-surface and off-surface points. Consequently, the weight function is given as:
\begin{equation}
w(\mathbf{X}; H_n) = \mu_w(\mathbf{X}).
\end{equation}

As the kernel implies neighboring points own similar function values, close points of unprofitable points gain less weight than those of on-surface points. 

\noindent\textbf{Exploration Function.~}
As the range of uncertainty function and weight functions lies in [0, 1], the exploration function is defined as the product of uncertainty function and weight function:
\begin{equation}
e(\mathbf{X}) = u(\mathbf{X})w(\mathbf{X}).
\end{equation}

The exploration function suggests the next point by finding the $\mathbf{x}$ with the highest function value. In this way, the suggested point has a higher possibility of lying in on-surface space, based on higher weights placed at on-surface points. Compared with using uncertainty function alone, the exploration function suggests points near the on-surface points, which avoids redundant exploration in unprofitable areas. 

	 

\section{Experiments and Results}
\subsection{Experimental Setting}

We set up our experiment on Baxter robot's 7 degrees of freedom arm. We only use the torque-force sensors on the end-effector to determine whether there is a collision with objects by measuring the force changes on the end-of-arm tooling during the tapping motion. A tapping tool is attached at the end of robot arm with fixed poses to touch the object. The tapping is conducted by the robot arm moving down at 3 \textit{cm} per step with 0.5 second pause. At each contact with objects, the end-effector pose within Cartesian coordinates by the arm kinematics is recorded as tapping points. The objects utilized include a wave shaped and slope shaped toy block with dimensions ($cm$) 16L $\times$ 6W $\times$ 11H  and 17L $\times$ 6W $\times$ 8H respectively. The objects are placed in a pre-defined 23L $\times$ 23W $cm$ searching area at the desk placed in front of Baxter robot where the kinematics of Baxter's left arm could be ideally solved. The searching space is highlighted within the grey transparent rectangle and the unprofitable tapping areas are uncovered desk area as shown in Fig. \ref{fig: result}(b).

\subsection{Results Analysis}

\begin{figure}%
\begin{subfigure}{.40\columnwidth}
\includegraphics[width=\columnwidth, height=0.11\textheight]{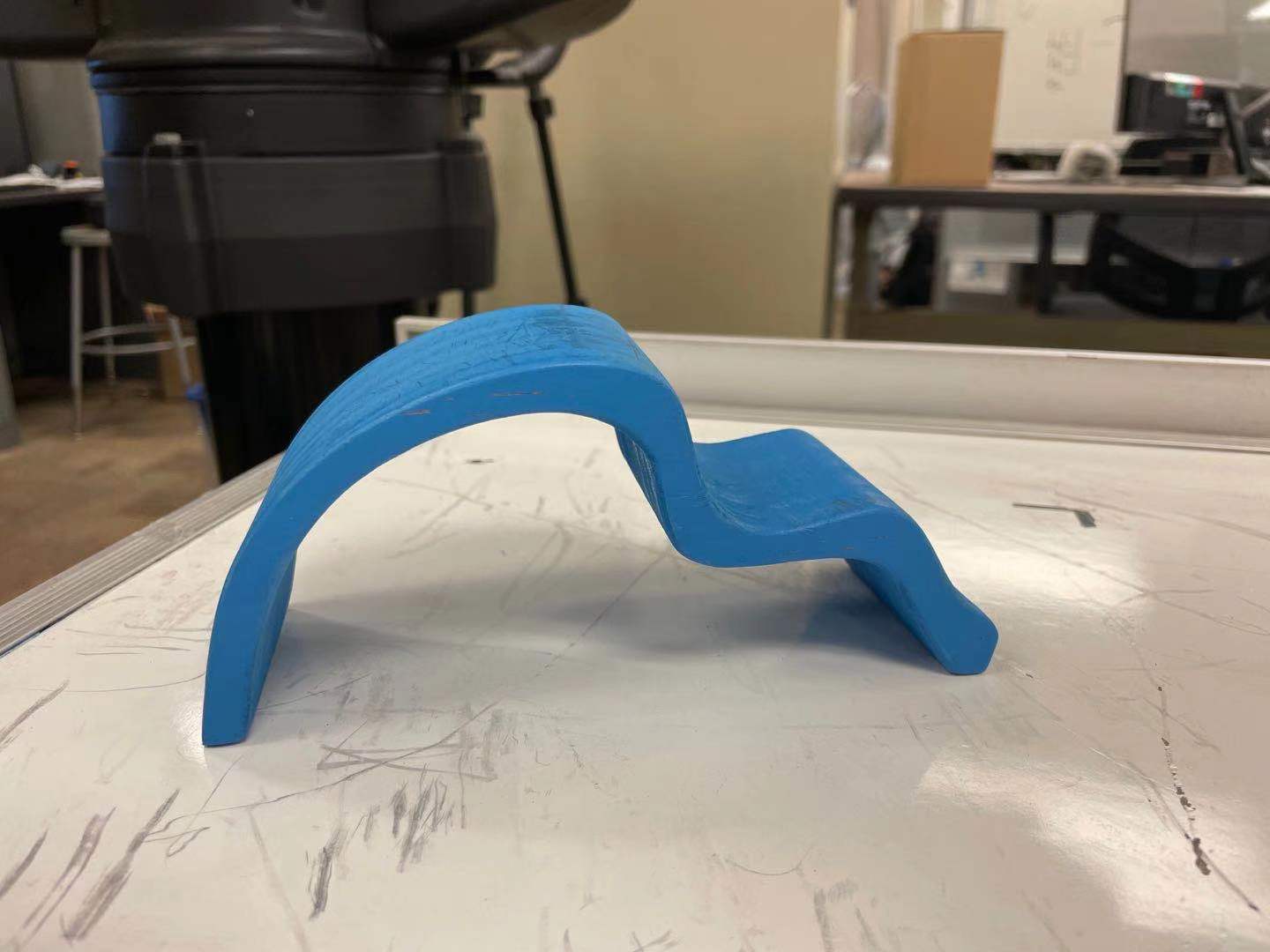}%
\caption{Object II}%
\label{subfiga}%
\end{subfigure}\hfill%
\begin{subfigure}{.59\columnwidth}
\includegraphics[width=\columnwidth, height=0.12\textheight]{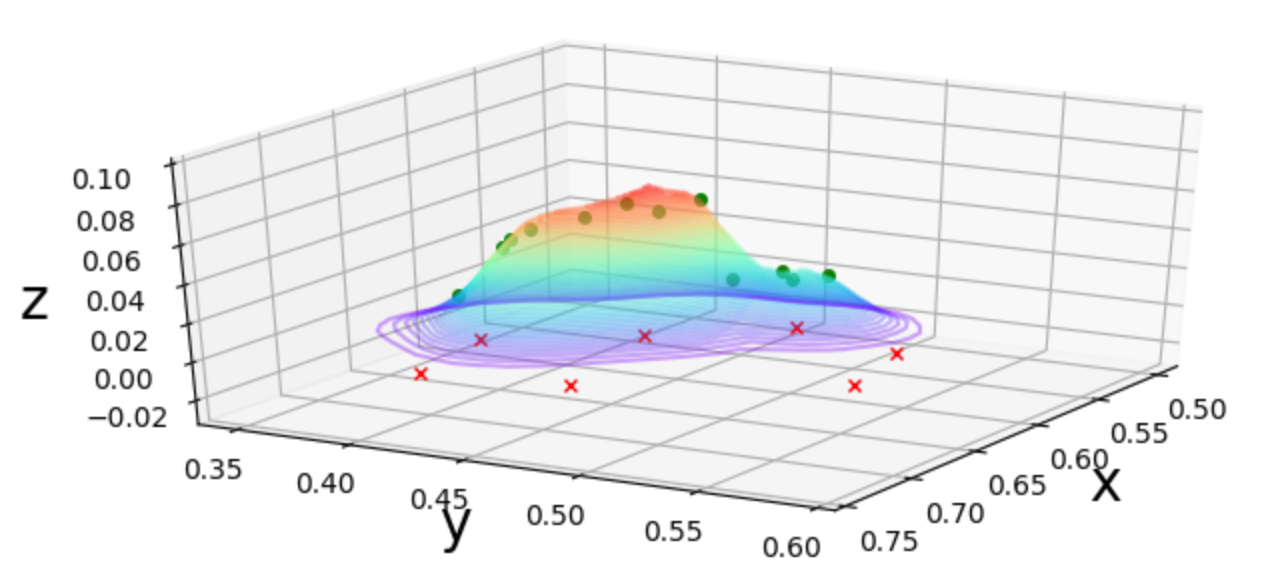}%
\caption{GP model}%
\label{subfigb}%
\end{subfigure}\hfill%
\caption{GP surface estimation of object II. \textit{Please zoom-in on (b) for details.}}
\label{fig: object2}
\end{figure}

Fig. \ref{fig: result} shows the results of estimating surface of object I from 14 iterations, and Fig. \ref{fig: object2} shows the results of surface reconstruction of Object II from 19 taps. In Fig. \ref{fig: result}, we can see that the suggestion function tends to find the next point adjacent to known surface positions as off-surface positions are placed with relatively low weights by the weight function. Therefore, uncertainties within on-surface areas are minimized in priority, and exploration regions are expanded to unknown areas only when a neighboring point is suggested and proved to be on object surface.

As mentioned in our problem statement, given a searching space without the object region, there exits certain areas that objects are not occupied which we call unprofitable space. Because every tapping motion is expensive, unprofitable tapping is required to be minimized while on-surface tapping needs to be maximized to enhance efficiency and accuracy. 

We compared our weighted exploration-based approach with Yi's\cite{7759723} uncertainty-based method in the performance of active surface reconstruction given no prior knowledge of object region.
\begin{figure}[t]
\centering 
\begin{subfigure}{0.33\columnwidth}
  \includegraphics[width=\columnwidth,height=0.10\textheight]{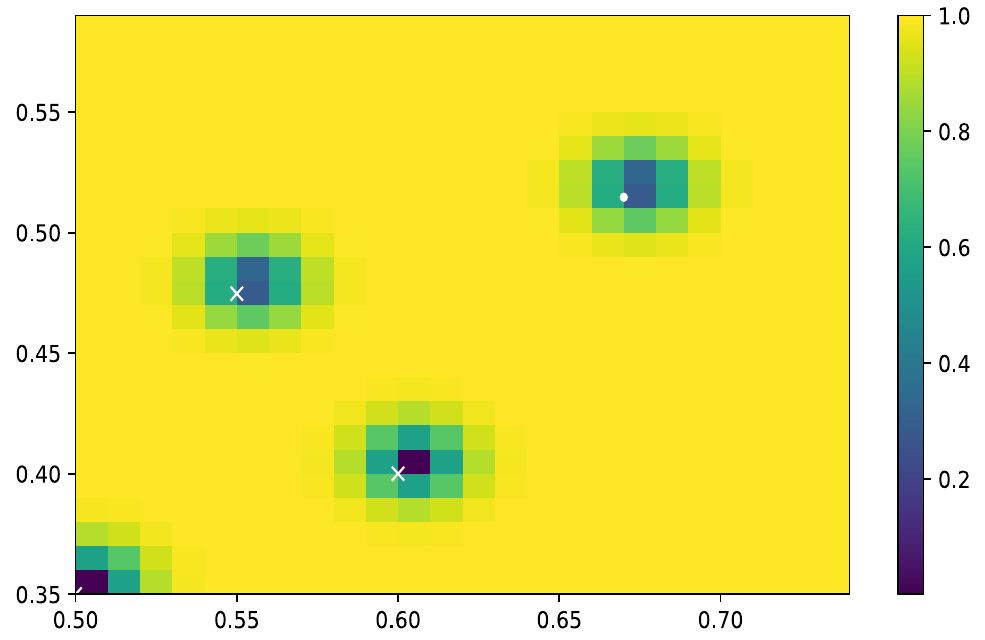}\hspace{0.1em}%
  \label{fig:1}
\end{subfigure}\hfill%
\begin{subfigure}{0.33\columnwidth}
  \includegraphics[width=\columnwidth,height=0.10\textheight]{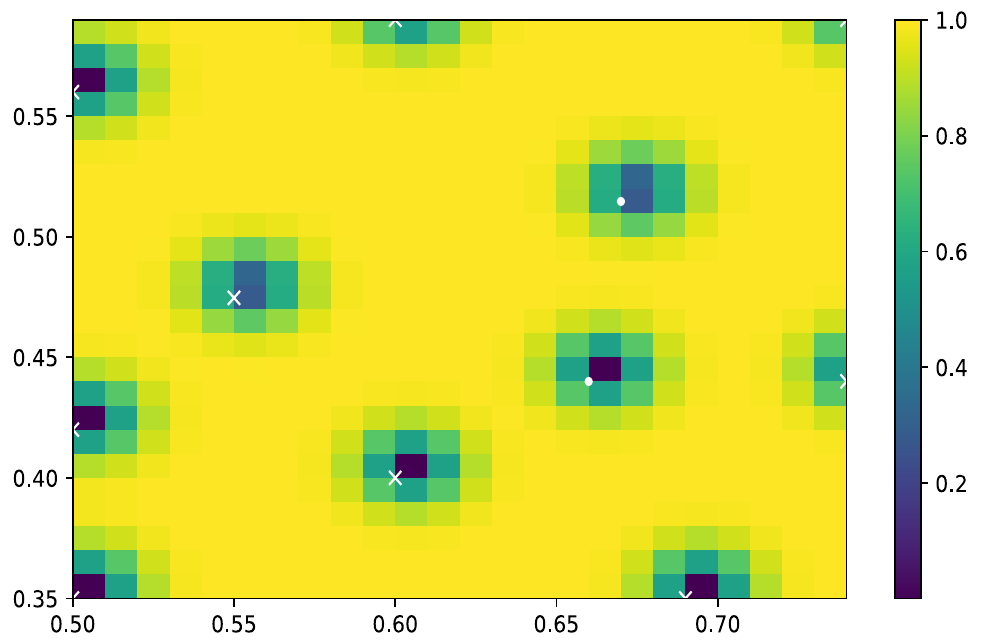}\hspace{0.1em}%
  \label{fig:3}
  \end{subfigure}\hfill%
 \begin{subfigure}{0.33\columnwidth}
  \includegraphics[width=\columnwidth,height=0.10\textheight]{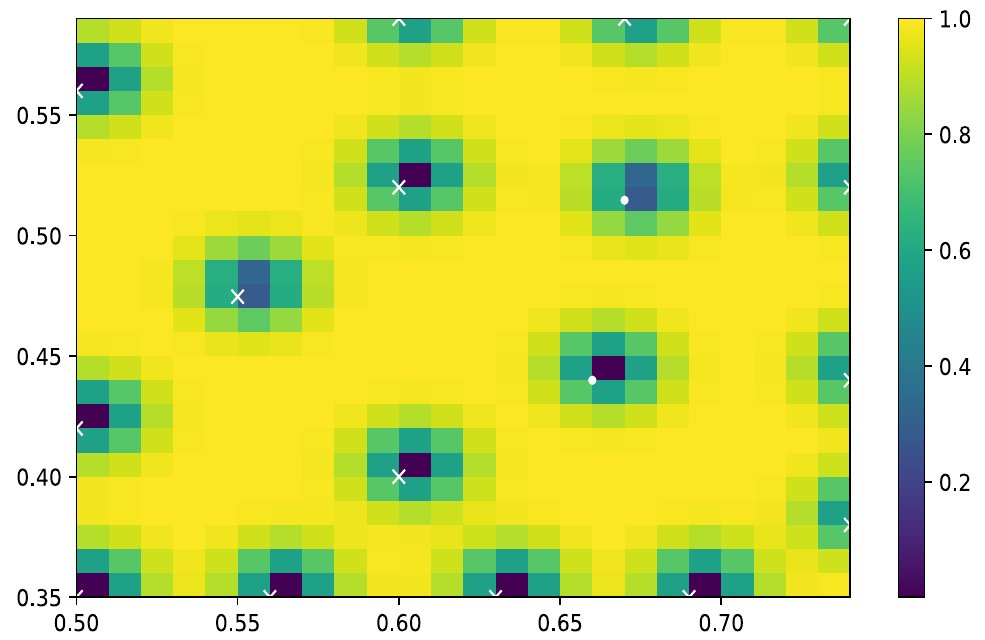}\hspace{0.1em}%
  \label{fig:5}
\end{subfigure}
\medskip
\begin{subfigure}{0.33\columnwidth}
  \includegraphics[width=\columnwidth,height=0.10\textheight]{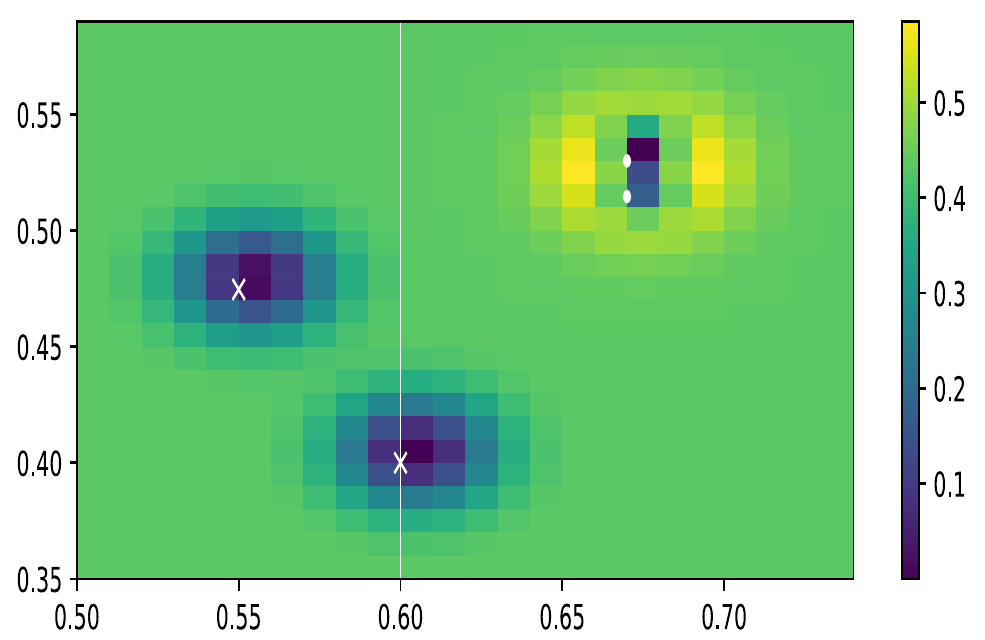}\hspace{0.1em}%
  \caption{Iteration 1}
  \label{fig:1}
\end{subfigure}\hfill%
\begin{subfigure}{0.33\columnwidth}
  \includegraphics[width=\columnwidth,height=0.10\textheight]{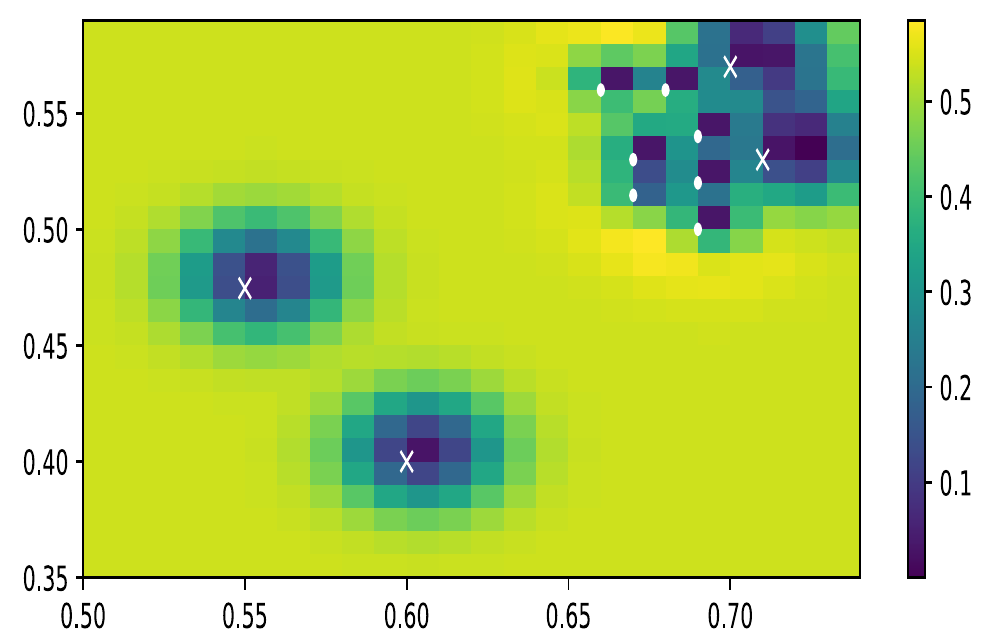}\hspace{0.1em}%
  \caption{Iteration 10}
  \label{fig:2}
\end{subfigure}\hfill%
 \begin{subfigure}{0.33\columnwidth}
  \includegraphics[width=\columnwidth,height=0.10\textheight]{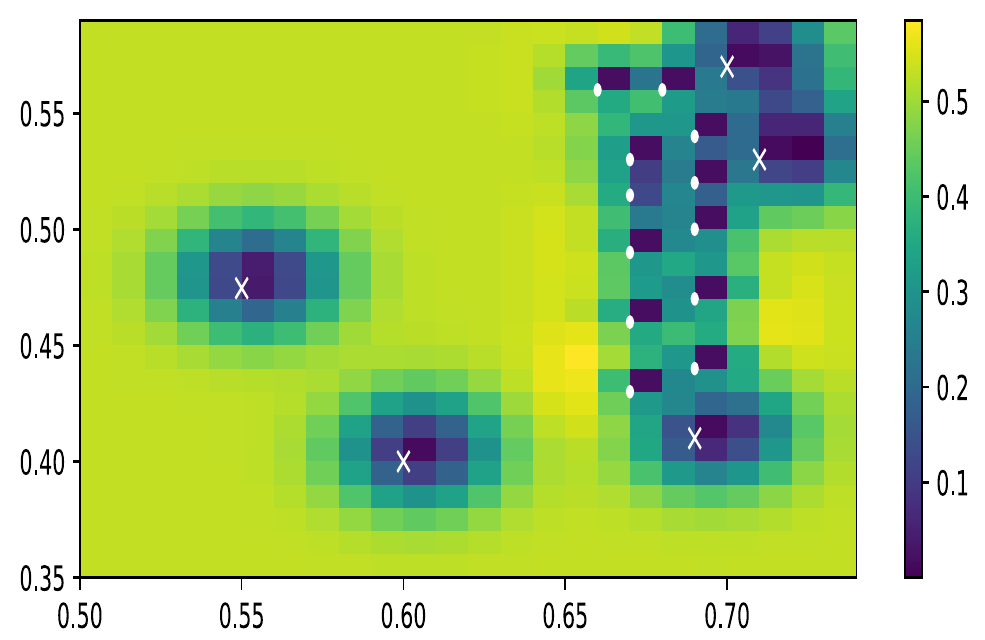}\hspace{0.1em}%
  \caption{Iteration 16}
  \label{fig:5}
\end{subfigure}\hfill%
\vspace{-2mm}
\caption{The result of three iterations within the exploration and reconstruction process. The top row shows the result of uncertainty metric-based  method, and the second row shows the results using our proposed method. The exploration procedure is conducted with Object I placed on the right side of searching space. \textit{Please zoom-in for details.}}
\label{fig:topview}
\end{figure}
\begin{figure}[t]
    \centering 
\begin{subfigure}{0.33\columnwidth}
  \includegraphics[width=\columnwidth, height=0.10\textheight]{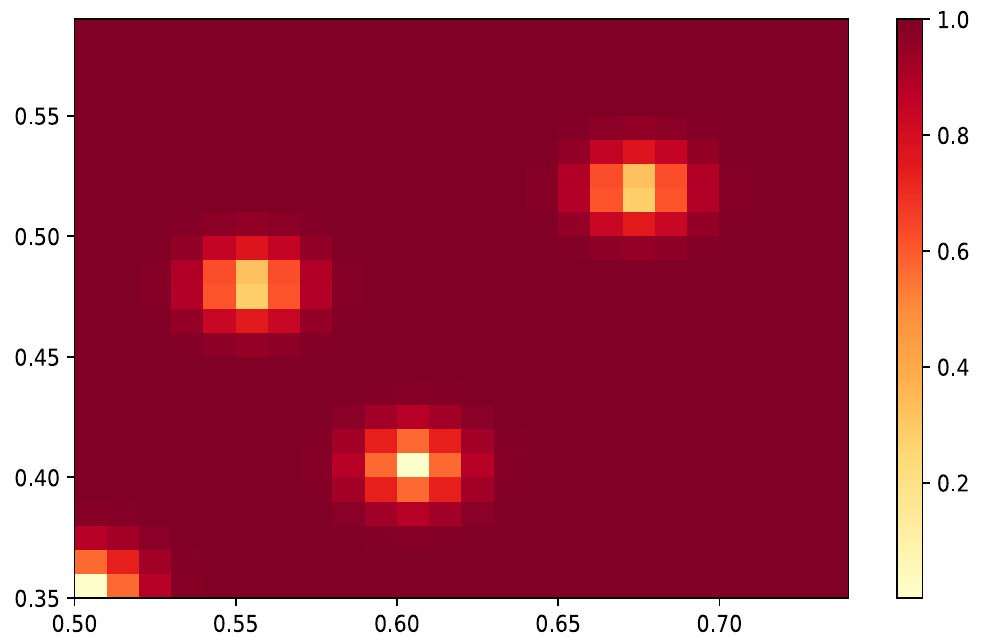}\hspace{0.1em}%
  \label{fig:1}
\end{subfigure}\hfill%
\begin{subfigure}{0.33\columnwidth}
  \includegraphics[width=\columnwidth, height=0.10\textheight]{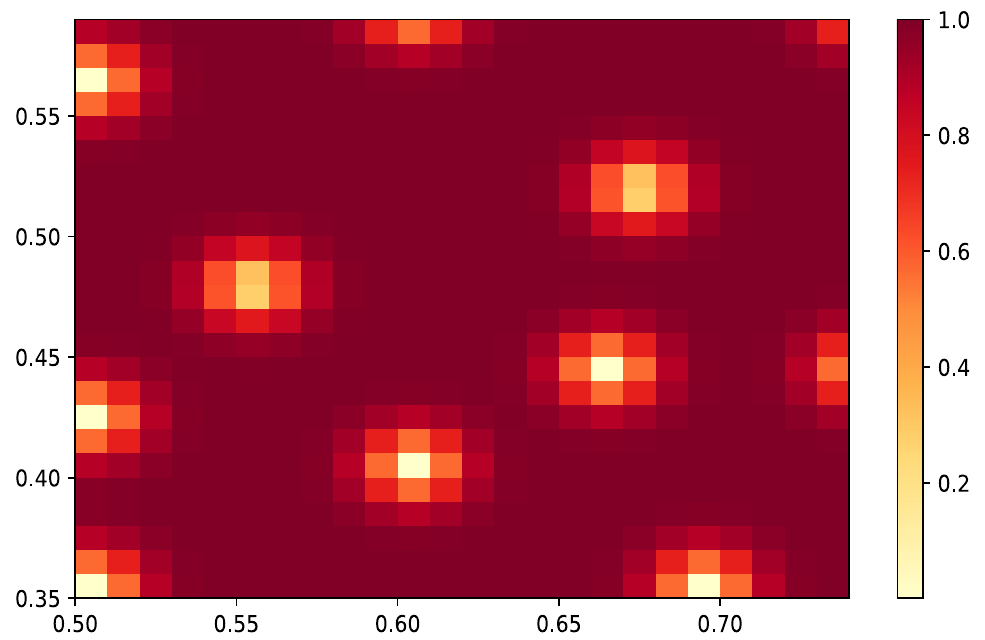}\hspace{0.1em}%
  \label{fig:3}
  \end{subfigure}\hfill%
 \begin{subfigure}{0.33\columnwidth}
  \includegraphics[width=\columnwidth, height=0.10\textheight]{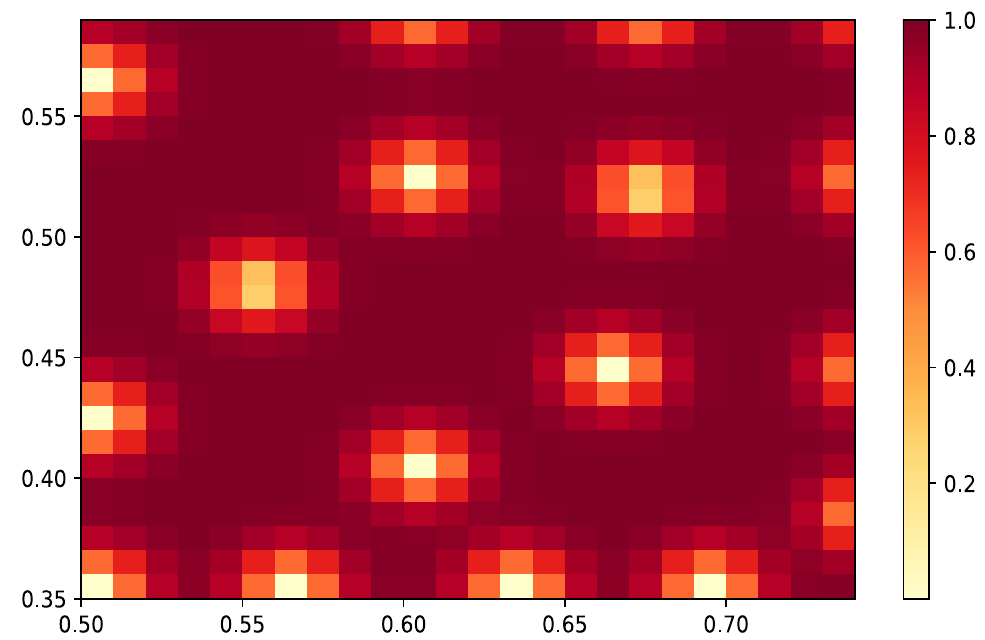}\hspace{0.1em}%
  \label{fig:5}
\end{subfigure}\hfill%
\medskip
\begin{subfigure}{0.33\columnwidth}
  \includegraphics[width=\columnwidth, height=0.10\textheight]{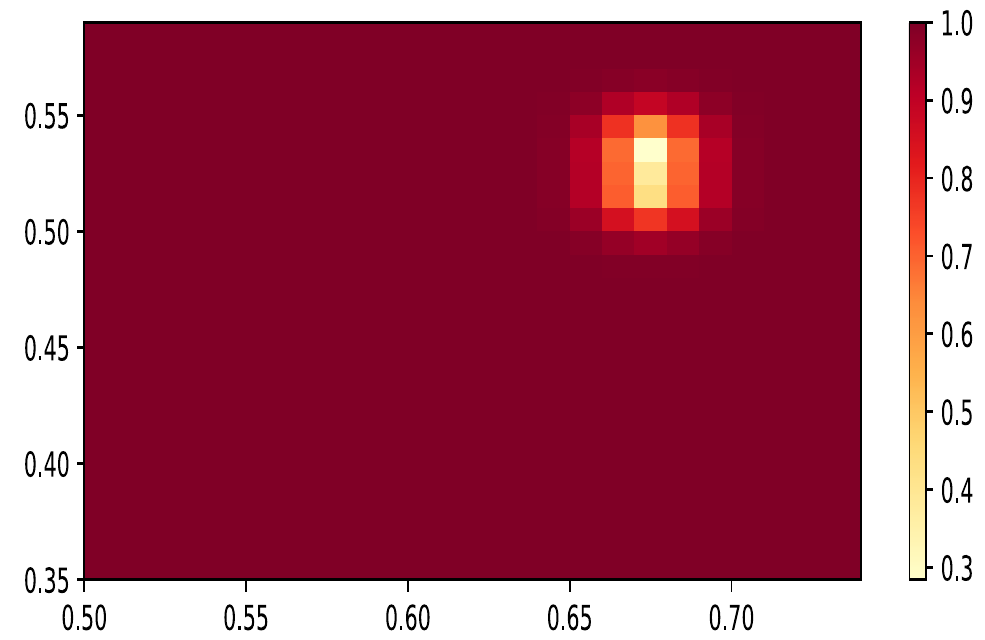}\hspace{0.1em}%
  \caption{Iteration 1}
  \label{fig:1}
\end{subfigure}\hfill%
\begin{subfigure}{0.33\columnwidth}
  \includegraphics[width=\columnwidth, height=0.10\textheight]{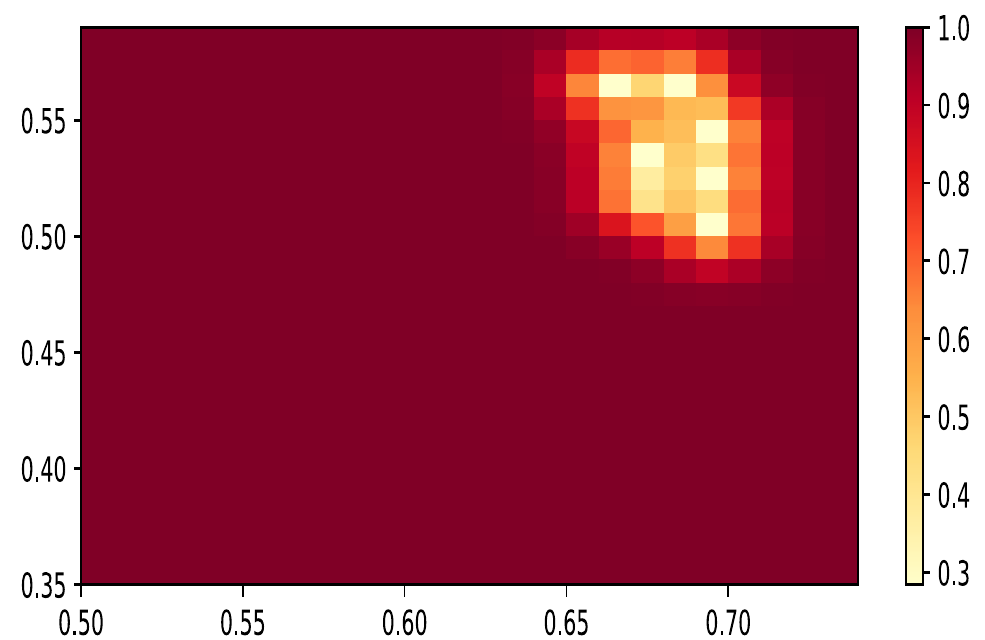}\hspace{0.1em}%
  \caption{Iteration 10}
  \label{fig:3}
  \end{subfigure}\hfill%
 \begin{subfigure}{0.33\columnwidth}
  \includegraphics[width=\columnwidth, height=0.10\textheight]{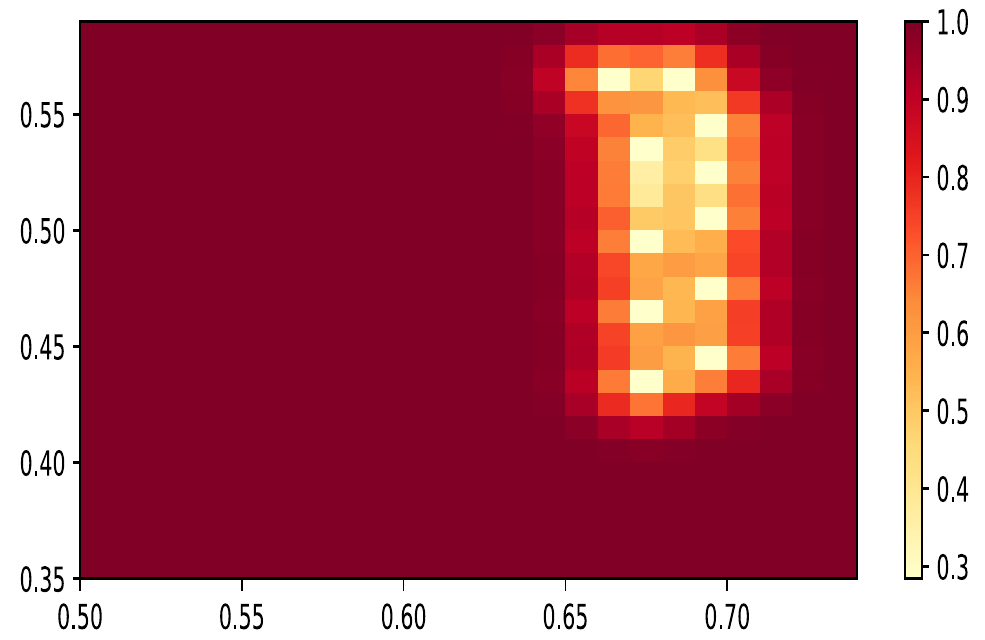}\hspace{0.1em}%
  \caption{Iteration 16}
  \label{fig:5}
\end{subfigure}\hfill%
\vspace{-2mm}
\caption{Comparison of minimized uncertainties distribution result within searching space by uncertainty function (top row) and exploration function (second row). The procedure is conducted with Object I placed on the right side of searching space.}
\vspace{-5mm}
\label{fig:uncer}
\end{figure}
Fig. \ref{fig:topview} shows the result of selected iterations within the surface reconstruction process for object I placed in its searching place as in Fig. \ref{fig: result}(b). The `$\bullet$' in each figure indicates valid taps on object surface while `$\times$' encodes unprofitable taps on the desk. The background colormap indicates the probability distribution of selecting next tapping position. Areas with lighter color indicate a higher probability of object existence while darker areas indicate a higher probability of unprofitable tapping. The first row is the results of uncertainty function-based GP surface reconstruction. Within a total number of 17 taps, only 2 taps happened on object surface and 15 taps happened in unprofitable areas. The reason is that within the searching space, uncertainties not only exists on object surface but also prevails within unprofitable space. 

With the goal of only minimizing the uncertainties in a searching space which is considerably wider than the object region, suggested tapping positions by the uncertainty function are distributed evenly within the entire space. Shown by the figures in the first row, the entire area was explored. Accordingly, on-surface area which is essential was not emphasized and an unignorable number of unprofitable taps was executed. The second row shows the results of our approach. Compared with the uncertainty-based method, only 5 taps were within unprofitable areas, indicating a 59\% improvement in the percent of effective taps among 17 taps in total. More notably, the unprofitable taps mainly took place around the object contour, and valid taps were centered on the object coverage areas, as evident by the last figure. In reliance on our proposed exploration function, areas in close proximity with valid taps possess higher probability of embodying the target object, which can be more explicitly observed from the first three figures. 

In the first row of Fig. \ref{fig:uncer}, it is shown that the uncertainty-based tapping result minimizes the uncertainties of the whole searching region. Our approach, shown in the second row, selectively minimizes the uncertainty distributed within object coverage area and evades the unprofitable space.

\section{Conclusion}
In this paper, we proposed an active tapping method to estimate object surface distribution based on Gaussian Process Regression (GPR) model. Different from the existing literature that requires prior information of object region, we implemented our study with unknown objects in a searching space consisting of on-surface contact points and off-surface contact points. As the off-surface tapping is considered as expensive, we propose an active tapping method to guide the tapping positions intelligently. Our method incorporates minimization of uncertainties existing in searching area with a priority of object coverage area, by a weight function to estimate probability distribution of surface points. We validated our method on a Baxter robot arm executing tapping motions with a tapping tool attached at the end-effector on objects. The results showed that the object surface can be estimated with a small number of tapping trails. 

\bibliographystyle{IEEEtran}
\bibliography{reference}

\end{document}